# One-Index Vector Quantization Based Adversarial Attack on Image Classification

Haiju Fan [a][b], Xiaona Qin [a][b], Shuang Chen [c], Hubert P. H. Shum [c][†] and Ming Li [a][b]

[a] College of Computer and Information Engineering, Henan Normal University, Henan 453007, China
[b] Key Laboratory of Artificial Intelligence and Personalized Learning in Education of Henan Province, Xinxiang 453007, China
[c] Department of Computer Science, Durham University, Durham, UK      [†] Corresponding Author

A B S T R A C T

To improve storage and transmission, images are generally compressed. Vector quantization (VQ) is a popular compression method as it has a high compression ratio that suppresses other compression techniques. Despite this, existing adversarial attack methods on image classification are mostly performed in the pixel domain with few exceptions in the compressed domain, making them less applicable in real-world scenarios. In this paper, we propose a novel one-index attack method in the VQ domain to generate adversarial images by a differential evolution algorithm, successfully resulting in image misclassification in victim models. The one-index attack method modifies a single index in the compressed data stream so that the decompressed image is misclassified. It only needs to modify a single VQ index to realize an attack, which limits the number of perturbed indexes. The proposed method belongs to a semi-black-box attack, which is more in line with the actual attack scenario. We apply our method to attack three popular image classification models, i.e., Resnet, NIN, and VGG16. On average, 55.9% and 77.4% of the images in CIFAR-10 and Fashion MNIST, respectively, are successfully attacked, with a high level of misclassification confidence and a low level of image perturbation.

## 1. Introduction

Many images are transmitted over public channels in computer networks through techniques such as encryption, compression and watermarking [1]. However, these images are vulnerable to attack. An attacker may capture data and retransmit the tampered version in the common channel, resulting in great security risks. In recent years, deep learning has been prevalent with many fruitful results in image classification. While neural network classification models have surpassed humans in recognition speed [2], a leak has been found in numerous studies that these models are vulnerable to artificial attacks [11]. In particular, image adversarial attacks have shown to be highly effective in fooling these networks, i.e. misleading the classification with small perturbations to the input images, leading to network security risks [12].

As a relatively emerging field, adversarial attacks have been extensively investigated with many attack methods derived, such as the Fast Gradient Sign Method (FGSM) [14], Universal Adversarial Perturbation (UAP) [25], Jacobian-based Saliency Map Attack (JSMA) [26]. In the above attack methods, the number of pixels modified is high. In the extreme case, the one-pixel attack [21] was proposed, where only one pixel can be modified to achieve an adversarial attack without restricting the strength of modification. These methods are generally applied to the pixel domain, which is not a popular domain where images are stored and transmitted in the real world.

In reality, most images are stored and transmitted in compressed form instead of uncompressed form. Therefore, methods attacking the compression domain have a much wider range of application scenarios. Most of the existing adversarial attack methods directly added perturbations to the redundant parts. For example, in the UAP method [23], the attacker captures an image transmitted within common channels and adds pixel-level perturbations to natural images to mislead the model output. However, compressed images only retain important information and remove redundant information, making it more difficult for attacks to be performed. In fact, it has been shown that JPEG compression effectively can be considered as a defence method against adversarial attacks by reducing adversarial noise [3]. This explains a lack of research in adversarial attacks focusing on the compressed domain.

In this paper, we propose a one-index adversarial attack method for the vector quantization domain under the semi-black-box setting. The one-index attack method modifies one VQ index element in the compressed data stream of the vector quantization domain to make the decoded image misclassified, which is comparable to one-pixel attacks in the pixel domain. Vector quantization [6] is a popular lossy compression technique, which has a high compression ratio that cannot be replaced by other technologies and has been widely used for many applications, such as JPEG compression [4] and Discrete Cosine Transform compression (DCT) [5]. Therefore, the application scenario of vector quantization domain-based attack methods is much broader. Finally, the semi-black-box attack setup is more in line with the actual attack scenarios, of which the attacker does not know the network model of structure or parameters [7,8], as opposed to the white box attack [9] where the attacker utilizes known information about the target model and can easily generate corresponding adversarial images [10]. This means that the available information is the VQ index and output probability label. The key idea of our method is to gradually decrease true category probability by modifying one VQ index using a differential evolution algorithm. In particular, we propose a codeword sorting algorithm that establishes correlations between neighboring indexes so that one-index perturbations can be optimized using differential evolutionary algorithms to improve

―――――――――
† Corresponding Author. Email: 121064@htu.edu.cn (H. Fan), 2108283082@stu.htu.edu.cn (X. Qin), shuang.chen@durham.ac.uk (S. Chen), hubert.shum@durham.ac.uk (H. P. H. Shum), liming@htu.edu.cn (M. Li)

optimization efficiency in the solution space.

The proposed one-index attack method is validated using three common network models (Resnet, NIN, VGG16) and image datasets (CIFAR-10, Fashion MNIST). We show that any input image from any source class can be perturbed to be misclassified with a 55.9% success rate and a 77.4% confidence on average, while only modifying a single VQ index. We also conduct ablation experiments for differential evolution and codeword sorting algorithms, showcasing an improved attack performance.

## 2. Related works

In recent years, the security of deep neural networks has been of great concern due to the weak interpretability of hidden layers. In computer vision, adversarial attack methods add small amounts of noise to natural images. The noise is imperceptible to human eyes and does not affect human recognition. However, it confuses the machine learning model to make a wrong decision [11]-[13]. According to the methodological basis, adversarial attack methods can be divided based on gradient, optimization, boundary, attention, and salience map.

Several works explored the gradient-based adversarial attack method. The Fast Gradient Sign Method (FGSM) considered reverse thinking based on the training network to attack a network model and mislead the model output [14]. It increased the loss function value in the gradient opposite direction and quickly generated adversarial images, although it had a low success rate. Iterative-FGSM (I-FGSM) imported step size for iteration on the FGSM method, improving the attack success rate but reducing the transferability [15]. To improve the attack rate, Projected Gradient Descent (PGD) was proposed as a multi-step iteration of FGSM [16]. Zeroed Order Optimization (ZOO) used Newton's method to iteratively obtain the best adversarial examples by estimating the values of the first-order gradient and the second-order gradient [17]. To reduce the time of gradient estimation, the Autoencoder-based Zeroth Order Optimization Method (AutoZOOM) [18] used an adaptive stochastic gradient estimation strategy and an autoencoder. Perturbation analysis of gradient-based adversarial attacks focused on designing novel methods for generating adversarial instances [19]. Jacobian-based Saliency Map Attack (JSMA) had a high success rate [26]. It identified key pixel positions based on gradient impact on classification and added perturbations. However, the generated adversarial samples are not migratable.

Optimization-based attacks optimized perturbations and found the adversarial perturbations to modify the number or strength as little as possible. The C&W method had superior attack performance and visual quality of adversarial images [20] but required a long time to search for the optimal parameters. The one-pixel attack method achieved image misclassification by modifying only one pixel [21].

Boundary-based adversarial attacks mainly investigated the decision classification boundaries of models. DeepFool found the minimum perturbation that changed the classification label of a sample by calculating the distance between the classification boundary and the sample [22]. However, it needed to modify more pixels. To improve the migration of perturbations, Universal Adversarial Perturbation (UAP) generated adversarial perturbations with strong generalization capability by calculating the shortest distance between the original sample and the model boundary [23]. However, the computation time for universal adversarial perturbation is long.

Some existing works transferred the model's attention to facilitate attack. Gradient-weighted Class Activation Mapping (Grad-CAM) modified the heat map regions for classification, resulting in misclassification [24]. Attention-Attack made the feature maps of the L-layer of images and the L-layer of target categories as similar as possible to transfer the model's attention and mislead the model [25].

The aforementioned attack methods are applied on the pixel domain and have a more limited application. Our proposed method is similar to one-pixel attacks [21] in the sense that the modification number is limited to one. While one-pixel attack is applied on the pixel domain, ours is on the vector quantization domain, which has a much broader application scenario. Also, the one-pixel attack method attacks the whole image pixels, while our method attacks the image compressed form.

## 3. Methodology

In this section, first, we propose the one-index attack algorithm, which modifies an index without restricting modification strength, to deceive the classification model. Then, to improve the attack success rate, we propose the differential evolutionary algorithm to optimize the one-index perturbation generation by generation and find the optimal perturbation that makes images misclassified. Finally, considering the evolutionary property of the differential evolutionary algorithm, when finding certain one-index perturbations with a successful attack, it tends to continue to search for a better perturbation in its vicinity, so we need to establish the correlation between the codeword index and vector. Thus, we propose the codeword sorting algorithm based on principal component analysis.

The proposed one-index adversarial attack method is not related to the network model structure. It realizes attack by decreasing true category probability labels instead of needing to know information such as parameters, which is more in accord with realistic attack scenarios. Our method is for compressed domain images, where the redundant information is removed when the image is compressed. There is almost no redundant information in the compressed image. Therefore, it is more difficult to attack the compressed domain than the pixel domain. Since the image is transmitted in compressed form in cyberspace, the adversarial attack method based on the compressed domain has a broader application scenario. The target of the proposed attack method is the image-compressed data stream, which has practical significance.

*3.1. Attack Scenarios and Notations*

In this paper, we employ a semi-black-box setting [7] where the attacker has no access to the inner information of the model, such as the gradient and parameters, but only the probability label. This aligns well with real-world scenarios, in which an attacker may query the victim model but does not have the model implementation details. See Table 1 for important notations.

*3.2. One-Index Attack in the Vector Quantization Domain*

Our method attacks directly on the compressed data stream instead of attacking the decompressed image, which is in line with the actual attack situation and has wide application scenarios. The proposed one-index attack method based on the vector quantization domain makes the decoded image misclassified by modifying one VQ index. In [21] attack method, it modifies multiple pixels and it is easy to find out that the image has been tampered with. Therefore, we limit the number of indexes to be modified.

We propose a one-index adversarial attack method targeting the vector quantization domain. Our method modifies only one VQ index to attack, and the VQ index's range is $[1, L]$, where L is the codebook length. The one VQ index perturbation $e(I)$ of the image $I$ is defined as shown in Eq. (1), where $x$ and $y$ respectively represent the $x$ and $y$ coordinates of the VQ index; and $r$, $g$, $b$ represent the attacked VQ index value at the corresponding position in the RGB layer:

**Table 1** Important notations and their descriptions.

| Notation | Description |
| --- | --- |
| $e(I)$, $F$ | Perturbation, model |
| $x$, $y$ | The location of perturbation |
| $r$, $g$, $b$ | The value of perturbation |
| $iter$ | Maximum number of iterations |

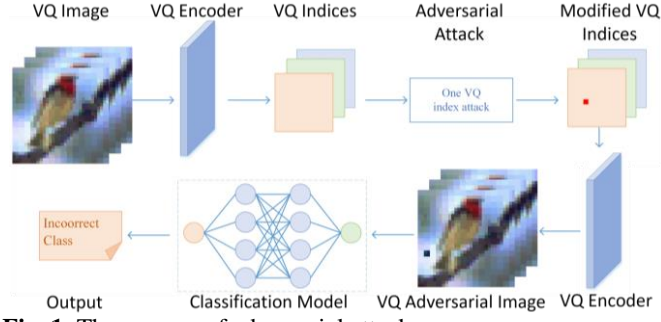

**Fig. 1.** The process of adversarial attack.

$$e(I) = \{x, y, r, g, b\} \qquad (1)$$

where $x \in [1, s]$, $y \in [1, t]$, $r, g, b \in [1, L]$ and $s, t$ represents the size of the VQ index matrix.

The method's objective is to minimize the true category probability, as shown in Eq. (2), where $F_t$ is the true category probability obtained by the model $F$ and $\|e(I)\|$ represents the modification VQ index quantity:

$$\min_{e(I)^*} F_t(I + e(I)) \text{ subjected to } \|e(I)\| = 1 \qquad (2)$$

The method modifies one VQ index to generate the adversarial images, which results in the model outputting an incorrect category, as shown in Fig. 1. The sender feeds images into the VQ encoder and obtains VQ indexes for transmission within the common channel. The attacker intercepts and tampers the compressed data stream – VQ indexes. The receiver gets the modified VQ index to reconstruct images from the decoder. The reconstructed image will make neural network models output a wrong category. The specific steps of one VQ index attack method are as follows:

1. Determining the modification location according to the x, y of one index perturbation $e(I)$.

2. Replacing the VQ index value at the corresponding position with the $r$, $g$, $b$ values of the perturbation $e(I)$ and obtain the attacked VQ index matrix.

3. Decoding the attacked VQ index to get the attacked image. If it can make the model output an incorrect category, the adversarial image is generated successfully.

**Algorithm 1** shows the one VQ index attack method, where $l_t$ is the true category, $l_f$ is the predicted category. We describe one VQ index attack problem as an optimization problem and optimize the one VQ index perturbation by differential evolution algorithm. The Differential Evolutionary (DE) [28] is a kind of intelligent optimization method that evolves populations by optimizing the fitness function value. Compared to the genetic algorithm [30], it maintains population diversity during the iterative process and jumps out of the local optimum easily. Thus, it can quickly find a global optimum in the solution space.

---

**Algorithm 1:** The one VQ index attack process

**Input:** VQ index $v$ and one index perturbation $e(I)$
**Output:** Attacked VQ index $v'$ and adversarial image $I'$
1. $e(I) = \{x, y, r, g, b\}$
2. **for** $k=1$ to 3 **do**
3.   **for** $i=1$ to $s$ **do**
4.     **for** $j=1$ to $t$ **do**
5.       $v'(i,j,k) = v(i,j,k)$
6.     **end**
7.   **end**
8. **end**
9. $v'(x,y,1) = r$, $v'(x,y,2) = g$, $v'(x,y,3) = b$
10. $v' \xrightarrow{\text{VQ decoder}} I'$
11. $I' \xrightarrow{\text{DNN model}} l_f$
12. **if** $l_f \neq l_t$
13.   Attack success
14. **else**
15.   Attack fails
16. **end**

---

### 3.3. Differential Evolutionary Perturbation Optimization

In particular, we consider the true category probability as fitness value. The optimization objective is to minimize fitness value and the constraint number of modified indexes is one. We set both the population size and the population iteration times to 50 after balancing the attack runtime and success rate. The specific steps of one VQ index perturbation optimization based on a differential evolutionary algorithm are as follows:

1. Generating a fixed number of perturbations as the initial population randomly, with each perturbation defined in Eq. (1).

2. Initializing the next generation population according to the mutation rules of the differential evolution algorithm as:

$$M_{i,k} = M_{\xi,k} + \varepsilon(M_{\varphi,k} - M_{\eta,k}) \qquad (3)$$

where $k$ is the current generation index; $M_{i,k}$ denotes the $i^{th}$ perturbation of the contemporary population; $M_{i,k+1}$ represents the $i^{th}$ perturbation of the next-generation population; $\xi$, $\varphi$, $\eta$ are random numbers, and $\varepsilon$ is the scale factor set to be 0.5.

3. Comparing the fitness value of adversarial images corresponding to child perturbations and parent perturbations. Taking the true probability labels as a measure of merit, the perturbation with the smallest probability label is selected to survive, gradually optimizing the population.

4. Repeating steps 2~3 to obtain the next generation population that is better than the current one. We repeat the algorithm to generate the next generation population, continue iterating until the maximum number of iteration times, and obtain the optimal population, resulting in the global optimal perturbation.

The differential evolution process is shown in **Algorithm 2**, where $p_s$ refers to the population size; $iter$ represents population iteration times.

---

**Algorithm 2:** DE algorithm

**Input:** The original image $I$
**Output:** The optimal perturbation $\mathcal{H}$
1. Initializing the population;
2. **for** $i=1$ to $p_s$ **do**
3.   **for** $j=1$ to 5 **do**
4.     $\mathcal{M}_{i,1}^j = \mathcal{M}_{min}^j + rand(0,1) * (\mathcal{M}_{max}^j - \mathcal{M}_{min}^j)$
5.   **end**
6. **end**
7. **for** $z=1$ to $iter$ **do**
8.   **for** $i=1$ to $p_s$ **do**
9.     **for** $j=1$ to 5 **do**
10.       *Mutation*;
11.       $\{\xi, \varphi, \eta\} = random(1, popsize)$
12.       $v_{i,z+1}^j = \mathcal{M}_{\xi,z}^j + \varepsilon(\mathcal{M}_{\varphi,z}^j - \mathcal{M}_{\eta,z}^j)$
13.     **end**
14.     *Selection*;
15.     **if** $F_t(I + v_{i,z+1}) < F_t(I + \mathcal{M}_{i,z})$
16.       $\mathcal{M}_{i,z+1} = v_{i,z+1}$
17.     **else**
18.       $\mathcal{M}_{i,z+1} = \mathcal{M}_{i,z}$
19.     **end**
20.   **end**
21. **end**
22. $i = argmin_{i=1}^{popsize} F_t(I + \mathcal{M}_{i,iter})$
23. $\mathcal{H} = \mathcal{M}_{i,iter}$

---

### 3.4. Codeword Sorting

In the differential evolution algorithm, if a certain perturbation makes the image misclassify successfully, it will tend to find the nearby perturbation and see whether it is better than that perturbation [28]. Thus, we need to establish the correlation between the codeword index and the corresponding codeword vector. To sort codewords in the codebook, we need to perform data dimensionality reduction. Common data dimensionality reduction methods include Principal Component Analysis (PCA), Independent Component Analysis (ICA), LinearDiscriminant Analysis (LDA), t-Stochastic Neighborhood Embedding (t-SNE), etc. PCA is the more basic linear dimensionality reduction method, so we use the PCA algorithm for codeword sorting.

Here, we aim to improve the correlation between adjacent codeword indexes to facilitate the effective convergence of the differential evolution algorithm. We first sort the codewords in the codebook by principal component analysis (PCA) before attacking [29]. Then, the column vectors corresponding to adjacent codeword indexes have a relatively strong correlation.

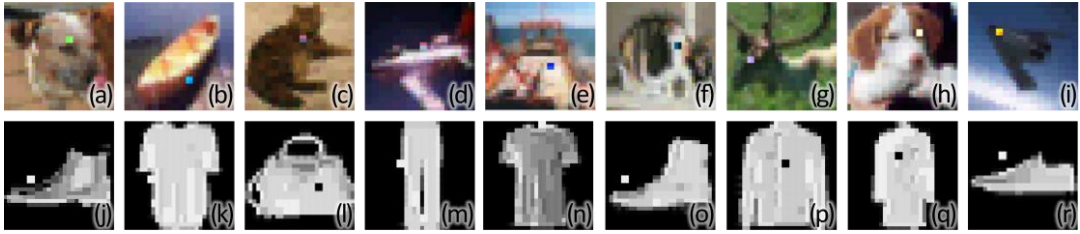

**Fig. 4.** Adversarial images of successful attacks on three network models on two datasets.

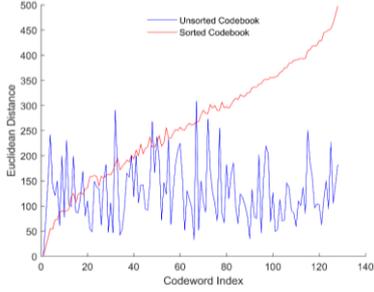

**Fig. 2** The part of $CB$ and $CB'$.

**Fig. 3.** The comparison of the unsorted and sorted codebook.

The specific steps of codeword sorting are as follows:

**Algorithm 3:** Codeword sorting algorithm
**Input:** The unsorted codebook $CB$
**Output:** The sorted codebook $CB'$
1. **for** $i$=1 to $L$ **do**
2.   **for** $j$=1 to $w*h$ **do**
3.     $a_i = \sum_{j=1}^{w*h} Y_{i,j}/(w*h)$
4.   **end**
5.   $p_i = Y_i - a_i$
6. **end**
7. $O = cov(p)$
8. $(v_{i=1}^L, \lambda_{i=1}^L) = eig(O)$
9. **for** $i$=1 to $L$ **do**
10.   $c_{i1} = v_1^T Y_i$
11. **end**

1. Pre-processing the codebook for standardization. Computing the average value $a_i$ of each codeword $Y_i$ in the codebook. Here, we get the matrix $P$ after preprocessing the codebook, which is defined as $P = \{p_1, p_2, ..., p_L\}$, where $p_i$ is calculated as:

$$p_i = Y_i - a_i \quad (4)$$

2. Computing the covariance matrix $O$ of the matrix $P$ according to Eqs. (5) and (6):

$$cov(Y_i, Y_j) = \frac{\sum_{q=1}^n (Y_{i,q} - a_i)(Y_{j,q} - a_j)}{n-1} \quad (5)$$

$$O = \begin{bmatrix} cov(Y_1, Y_1) & ... & cov(Y_1, Y_L) \\ cov(Y_2, Y_1) & ... & cov(Y_2, Y_L) \\ ... & ... & ... \\ cov(Y_L, Y_1) & ... & cov(Y_L, Y_L) \end{bmatrix} \quad (6)$$

3. Computing the eigenvalue $\lambda_1, \lambda_2, \lambda_3, ..., \lambda_L$ and eigenvectors of the covariance matrix $O$.

4. Sorting the column vectors in ascending order according to the first principal component to obtain the sorted codebook $CB'$.

As shown in Fig. 2, an unsorted codebook $CB$ displays disorder between codewords, while a sorted codebook $CB'$ shows a strong correlation between adjacent codewords. Taking the first codeword as an example, the comparison of Euclidean distance between the first codeword and other codewords in the unsorted codebook $CB$ and the sorted codebook $CB'$ is shown in Fig. 3. The Euclidean distance between two codewords is calculated according to Eq. (7): As shown in Fig. 3, the Euclidean distances of codeword indexes and their corresponding column vectors in unsorted codebooks are not correlated and non-regular, while the Euclidean distance between codewords in sorted codebook becomes larger as the index difference increases. The codeword index and corresponding column vector have a strong correlation. The codeword sorting algorithm based on principal component analysis is shown in **Algorithm 3**.

$$d(Y_i, Y_j) = \|Y_i - Y_j\| = \left[\sum_{n=1}^{w \times h} (y_{i,n} - y_{j,n})^2\right]^{1/2} \quad (7)$$

## 4. Experiments

In this section, we first evaluate the attack performance of the proposed method. Second, we compare the random search method, the DE search method (unsorted codebook) with our method for ablation experiments. Third, we analyze the change in the fitness value. Fourth, we analyze the quantitative relationship between the original category and the attacked category of adversarial images. Lastly, we compare the proposed scheme with other attack schemes.

### 4.1. Experiment Setup

**Datasets.** We evaluate our attack method using two widely used dataset for data recognition, CIFAR-10 dataset and the Fashion MNIST dataset. CIFAR-10 consists of 50,000 training images and 10,000 test images and Fashion MNIST consists of 60,000 training images and 10,000 test images with 10 categories.

**Classification model.** We evaluate our approach on Resnet, NIN and VGG16. These three models are popular for image classification and have very different network structures, which can test the transfer ability of the proposed method.

**Evaluation metrics.** The attack performance can be evaluated in the following main aspects: attack success rate and confidence. The success rate is defined as the percentage of adversarial images that were successfully attacked. Confidence is defined as the incorrect output category probability. The higher these metrics, the better the attack performance.

### 4.2. Attack Performance

To verify the effectiveness of the one-index adversarial attack method, we conduct experiments for the CIFAR-10 and Fashion MNIST datasets separately to further verify the feasibility of the method. Since the datasets have different categories, we need to train the classification models for this dataset separately. The accuracies of the three models for the CIFAR-10 and Fashion MNIST datasets are shown in Table 2.

We take 500 vector quantization images to test the attack success rate on the three network models (Resnet, NIN, VGG16) for two datasets. Table 3 shows the success rate and confidence of the one-index adversarial attack method on the three networks. As shown in Table 3, our method works for three network models with a 55.9% success rate and a 77.4% confidence on average on two image datasets. Among the three network models, the attack method has high robustness to the NIN model and the success rate and confidence level are higher. The method can misclassify not only a particular model but also other models.

Fig. 4 shows the adversarial images generated by the proposed adversarial attack method on three network classification models. As can be seen from Fig. 4, it is possible to modify one index to realize misclassification. It does not affect the classification results of human-discriminated images, while it causes the model to output incorrect categories with high confidence. Table 4 shows the related information of adversarial images in Fig. 4.

**Table 2** The accuracy on three different models.

| Dataset | Model | Accuracy |
|---|---|---|
| CIFAR-10 | Resnet | 86.97% |
| | NIN | 78.22% |
| | VGG16 | 79.75% |
| Fashion MNIST | Resnet | 89.39% |
| | NIN | 90.28% |
| | VGG16 | 91.28% |

**Table 3** The attack performances on three models.

| Dataset | Model | Success rate | Confidence |
|---|---|---|---|
| CIFAR-10 | Resnet | 44.8% | 80.27% |
| | NIN | 56.4% | 79.42% |
| | VGG16 | 52.8% | 74.64% |
| Fashion MNIST | Resnet | 93.6% | 92.4% |
| | NIN | 23.0% | 69.7% |
| | VGG16 | 64.8% | 67.7% |

**Table 4** The related information of adversarial images.

| Dataset | Model | ID | Original Category | Attacked Category | Confidence |
|---|---|---|---|---|---|
| CIFAR-10 | Resnet | a | Dog | Cat | 87.73% |
| | Resnet | b | Boat | Bird | 94.23% |
| | Resnet | c | Cat | Dog | 87.79% |
| | NIN | d | Airplane | Ship | 95.76% |
| | NIN | e | Ship | Truck | 79.38% |
| | NIN | f | Cat | Dog | 99.31% |
| | VGG16 | g | Deer | Bird | 74.32% |
| | VGG16 | h | Dog | Cat | 84.76% |
| | VGG16 | i | Airplane | Bird | 99.70% |
| Fashion MNIST | Resnet | j | Sneaker | Sandal | 93.25% |
| | Resnet | k | T-shirt | Shirt | 93.69% |
| | Resnet | l | Bag | Sandal | 83.83% |
| | NIN | m | Trouser | Dress | 66.09% |
| | NIN | n | T-shirt | Shirt | 75.46% |
| | NIN | o | Ankle boot | Sneaker | 70.64% |
| | VGG16 | p | Coat | Pullover | 75.46% |
| | VGG16 | q | Coat | Dress | 95.21% |
| | VGG16 | r | Ankle boot | Sandal | 96.77% |

*4.3. Codeword Sorting Effect on Population Evolution*

We compare the DE evolutionary process using a sorted codebook with an unsorted codebook. Firstly, we use the unsorted codebook generated by the LBG algorithm. Take an image as an example, the location and index distribution of perturbations in the initial, middle, and optimal populations are shown in Fig. 5(a) and Fig. 6(a).

As shown in Fig. 5(a) and Fig. 6(a), we can see that the position can converge to near a certain position, but the VQ index value cannot converge to a certain index value, which is almost evenly distributed in the solution space. One index perturbation contains position ($x$, $y$) and index value ($r$, $g$, $b$). If the algorithm finds a good perturbation, it prefers to search for a better perturbation in the vicinity of that perturbation. Since codewords in the codebook generated by the LBG algorithm are uncorrelated with each other, it will affect the superiority-seeking efficiency of the differential evolution algorithm.

Then, we test the same image to execute the differential evolution using the sorted codebook. The location distribution and index distribution using the sorted codebook in the initial, middle, and optimal populations are shown in Fig. 5(b) and Fig. 6(b). We intuitively perceive that the population slowly converges to the nearby areas of the optimal perturbation and gradually approaches the optimal perturbation, which verifies the importance of the codeword sorting algorithm. Comparing Fig. 5 and 6, the convergence rate is greater than that of the unsorted codebook. Therefore, the codeword sorting algorithm does improve the convergence speed of the population.

*4.4. Change in Fitness Values*

In this section, we take images from the CIFAR-10 dataset for example and test the variation of fitness values. In this paper, we describe one VQ index attack problem as the optimization problem, using the differential evolutionary algorithm to optimize one index perturbation, where the optimization objective is to minimize the probability label of the true category. We select randomly 20 correct category images on three network models (Resnet, NIN, VGG16) and carry out one-index attacks to test the change of fitness values with the number of population iterations. The fitness values on three models are shown in Fig. 7, where the fitness value is set to be the true category probability label for each image.

As can be seen from Fig. 7 in the early population iteration, the convergence speed is fast and the probability of true category decreases remarkably; in the middle population iteration, the convergence speed is slow and the fitness values descend modestly; in the late population iteration, it gradually converges to a stable value. The probability of true category decreases as the number of population iterations increases and the population evolutionary process is as expected. Therefore, the proposed one VQ index attack method based on the differential evolutionary algorithm can effectively decrease the probability of true category and achieve image misclassification.

*4.5. Ablation Study*

We take the CIFAR-10 dataset for example to conduct the ablation study. We compare our method with the random search attack method and DE search attack method (unsorted codebook) to evaluate whether the DE algorithm and codeword sorting algorithm help to improve the attack success rate and confidence. Random search attack method:

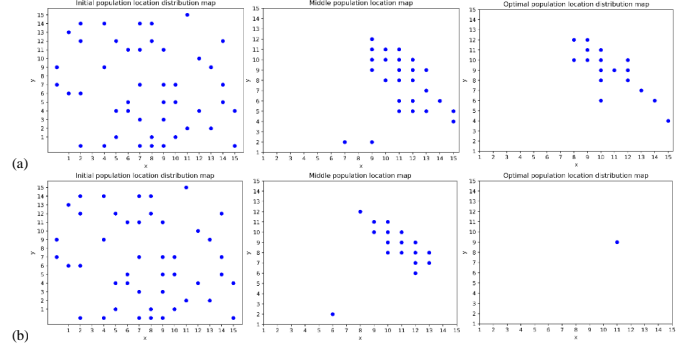

**Fig. 5.** Location distribution for (a) Unsorted & (b) sorted codebook.

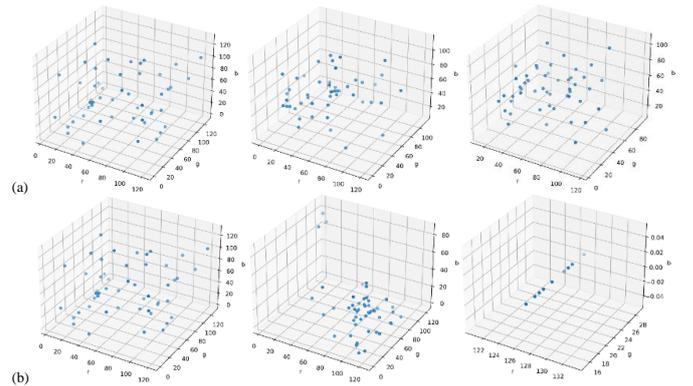

**Fig. 6.** Index distribution for (a) Unsorted & (b) sorted codebook.

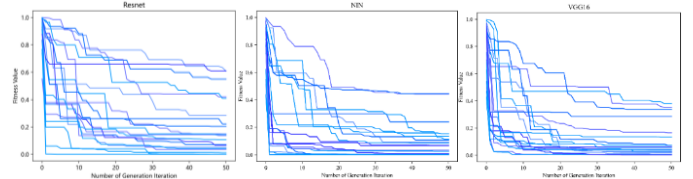

**Fig. 7.** The change of fitness value on three models.

For each image, the random search method repeats n times, which modifies one random VQ index each time. The confidence of adversarial images is set to the lowest true category probability among *n* attacks. DE search attack method (unsorted codebook): Not using the codeword sorting algorithm, DE search modifies one VQ index according to the differential evolution algorithm.

Here, we use the same number of evaluations (2500) for random and DE searches. From Method Ablation (Table 5 mid), our attack success rates on three models are higher than that of random search and DE search (unsorted codebook), showcasing that our method has the best attack performance. As shown in Average (bottom), even though random search is directionless, the attack method still has a 43.8% success rate on average. Thus, the vulnerable VQ indexes which can change the image label significantly are prevalent. Existing methods (top) achieve higher success rates. However, methods such as FGSM [14], One-pixel Attack [21], and Adversarial Steganography [27] are impractical in real-world applications due to their limited application scenarios. This will be clarified further in Section 4.7.

Our method uses a differential evolution algorithm to evolve one-index perturbations generation by generation, which is optimized in the direction of minimizing the true category probability of adversarial images. Thus, our attack success rate is 7.53% higher than random search on average. In the unsorted codebook, the adjacent codewords are not relevant, thus having a negative influence on the convergence rate. The DE search is 6.6% lower than our method on average. In addition, the average confidence of our method is higher than both these two methods.

*4.6. Adversarial Image Category Analysis*

We analyze the category of successful attack images on the CIFAR-10 dataset. Fig. 8 shows the heat maps of a successful attack, where $[i, j]$ indicates the number of images of category $i$ that have been attacked to category *j*. Vertical and horizontal indices denote the original and attacked categories respectively.

As can be seen from Fig. 8, two similar categories in general outline shape are prone to be successfully attacked between them. For example, cats and dogs are easily attacked successfully by each other. The three deep neural networks include convolutional layers and pooling layers. Some parts of images in two categories can get similar data points through the convolution kernel, and there are similar features. The same features can be obtained through the feature dimension reduction of the pooling layer.

Adding perturbations to such images will result in misclassifications. Cats are hardly attacked into cars as there are few similar overlapping features between the two categories. Therefore, the attack is more difficult. However, there are special boundary decision points. If the data point is modified, the image will be misclassified. The number of each category before and after the attack is shown in Fig. 9. Fig. 10 shows the difference for each category before and after the attack.

As shown in Fig. 8, using one VQ index attack method on the Resnet model, the NIN model, and the VGG16 model, every category of images is attacked successfully. Deers and ships are vulnerable to the other categories and the other categories are vulnerable to automobiles and birds. Among the 10 categories of images in the CIFAR-10 test dataset, certain categories are more robust than others, neither easily attack others nor easily be attacked by others. For these robust categories, a larger number of perturbations are required to successfully attack these images.

*4.7. Attack Performance Comparison*

The comparison of various attack methods is presented in Table 6, including the proposed attack method, FGSM method [14], one-pixel method [21], and adversarial steganography method [27]. Although [14] boasts the highest success rate and confidence among the other two methods, and it alters all image pixels. In contrast, [21] only changes one pixel value, but it still misleads the model output. Nevertheless, both methods add perturbations to image pixels in the spatial domain and are aimed at the image classification model. These methods have limited application scenarios since images are often transmitted in compressed form over network public channels. Compared to [14][21], which relies on the gradient, our method operates in a semi-black-box scenario, optimizing the perturbation based solely on output probabilities. Thus, our method is more versatile and applicable in scenarios where the gradient is inaccessible.

**Table 5** The attack performance (mid and bottom) against the impractical methods (top). **Bold** is the best, underline is the worst.

| | Method | Model | Success Rate | Confidence |
|---|---|---|---|---|
| Exiting Methods | FGSM [14] | NIN | 93.7% | 93% |
| | FGSM [14] | VGG16 | 90.9% | 90% |
| | One-pixel Attack [21] | NIN | 72.9% | 75% |
| | One-pixel Attack [21] | VGG16 | 63.5% | 65% |
| | Adversarial Steganography [27] | SRNet (J-UNIWARD) | 48.4% | - |
| | Adversarial Steganography [27] | SRNet (UERD) | 46.6% | - |
| Methods Ablation | Random Search | Resnet | 36.0% | **81.19%** |
| | DE Search | Resnet | 35.4% | 76.18% |
| | Proposed method | Resnet | **44.8%** | 80.27% |
| | Random Search | NIN | 48.0% | 73.54% |
| | DE Search | NIN | 50.4% | **81.78%** |
| | Proposed method | NIN | **56.4%** | 79.42% |
| | Random Search | VGG16 | 47.4% | 69.65% |
| | DE Search | VGG16 | 48.4% | **76.16%** |
| | Proposed method | VGG16 | **52.8%** | 74.64% |
| Average | Random Search | - | 43.8% | 74.8% |
| | DE Search | - | 44.7% | 78.0% |
| | Proposed method | - | **51.3%** | **78.1%** |

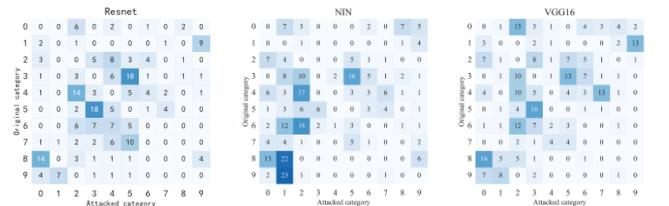

**Fig. 8.** Heat-maps for one-index attack on Resnet, NIN, and VGG16. The number 0 to 9 indicates the categories: airplane, car, bird, cat, deer, dog, frog, horse, ship, and truck.

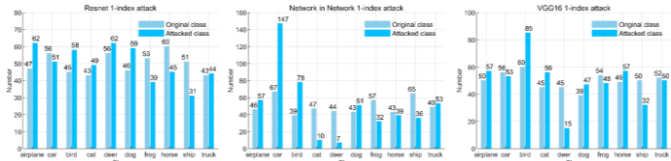

**Fig. 9.** Comparison of the number of original and attacked categories on the three networks.

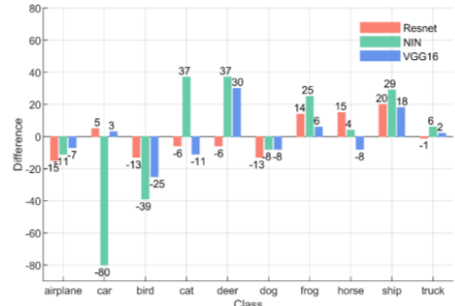

**Fig. 10** The difference between the number of original and attacked categories on the three networks.

**Table 6** The comparison of the related attack methods.

| Attack Method | Success Rate | Confidence | Network Model | Adversarial Target | Attack Domain |
|---|---|---|---|---|---|
| FGSM [14] | 93.7% | 93% | NIN | Classification | Spatial |
| FGSM [14] | 90.9% | 90% | VGG16 | Classification | Spatial |
| One-pixel Attack [21] | 72.9% | 75% | NIN | Classification | Spatial |
| One-pixel Attack [21] | 63.5% | 65% | VGG16 | Classification | Spatial |
| Adversarial Steganography [27] | 48.4% | - | SRNet (J-UNIWARD) | Steganalysis | JPEG |
| Adversarial Steganography [27] | 46.6% | - | SRNet (UERD) | Steganalysis | JPEG |
| Proposed method | 44.8% | 80% | Resnet | Classification | VQ |
| Proposed method | 56.4% | 79% | NIN | Classification | VQ |
| Proposed method | 52.8% | 75% | VGG16 | Classification | VQ |

In contrast, the adversarial steganography method [27] can be applied to the JPEG domain and targets the steganalysis model, which indicates that attacking images in the compressed domain is more difficult than in the spatial domain and the success rate is lower. The proposed attack method is applied in the vector quantization domain and specifically targets the image classification model. The original and adversarial images have a similar visual perception, striking a good balance between visual quality and attack success rate. Our method attacks the VQ index, offering insight into attacking other compression domain images.

## 5. Conclusion and Discussions

We present a novel one-index adversarial attack method grounded in vector quantization theory and differential evolution. The proposed theoretical framework offers an advancement in understanding adversarial attacks in the compressed domain.

Our work has both practical and theoretical significance. Practically, we show the feasibility of adversarial attacks in real-world scenarios where images are compressed. Theoretically, we provide a basis for further research into adversarial robustness in various domains, such as the VQ domain.

However, there are certain limitations. In the proposed method, the variation of one-index perturbation is limited to codewords. In future work, we can extend the codebook length by adding irrelevant codewords to expand the range of solution space to improve attack performance. In addition, we can explore the application of the diffusion model framework [31] in VQ attacks to improve attack performance, which has shown promising results in attacks in the pixel domain.

**Acknowledgement**

This work was supported by the Technology Attack Plan Project of Henan Province (222102210029), Key Program of the Higher Education Institutions of Henan Province (Grant No. 23A520009), and EPSRC NortHFutures (ref: EP/X031012/1).